%% file: Main.tex
\documentclass[10pt,twocolumn,letterpaper]{article}

\usepackage[workshop]{wacv}        

\usepackage{tcolorbox}
\usepackage{fontawesome5}
\usepackage{graphicx}
\usepackage{amsmath}
\usepackage{capt-of}
\usepackage{amssymb}
\usepackage{booktabs}
\usepackage{multirow}
\usepackage{xcolor}
\usepackage{array}
\usepackage{colortbl}

\definecolor{oursbg}{RGB}{226,242,228}

\usepackage[pagebackref,breaklinks,colorlinks]{hyperref}

\usepackage[capitalize]{cleveref}
\crefname{section}{Sec.}{Secs.}
\Crefname{section}{Section}{Sections}
\Crefname{table}{Table}{Tables}
\crefname{table}{Tab.}{Tabs.}

\def\wacvPaperID{40} 
\def\confName{WACV}
\def\confYear{2027}

\usepackage{pifont}\newcommand{\cmark}{\ding{51}}\newcommand{\xmark}{\ding{55}}
\newcommand{\ADD}[1]{{\color{black}{#1}}}
\newcommand{\TANI}[1]{{\color{black}{#1}}}

\begin{document}

\title{RS-JEPA: Remote Sensing Joint-Embedding Predictive Architecture\\with Scan-Path Views for Cross-Sensor Image Retrieval}
\author{
Tanishq Rachamalla$^{1}$ \qquad Srishti Kaushik$^{1}$ \qquad Muhammad Attique Khan$^{1,}$\\
\qquad Swalpa Kumar Roy$^{2}$\\[0.5em]
{\footnotesize $^{1}$Prince Mohammad Bin Fahd University, Al Khobar, Saudi Arabia \quad
$^{2}$Tezpur University, Assam, India}\\[0.3em]
{\tt\footnotesize tanishqrachamalla12@gmail.com, kaushiksrishti108@gmail.com, attique.khan@ieee.org,}\\
{\tt\footnotesize, swalpa@tezu.ernet.in}
}
\maketitle

\begin{abstract}
    \input{sec/0_abstract}
\end{abstract}
\vspace{-2mm}
\input{sec/1_introduction}

\input{sec/2_related_works}
\input{sec/3_methodology}

\input{sec/4_experiments}
\input{sec/5_results}

\input{sec/6_ablation}

\input{sec/7_conclusion}

{\small
\bibliographystyle{ieee_fullname}
\bibliography{egbib}
}

\end{document}

%% file: sec/0_abstract.tex
Cross-sensor remote sensing image retrieval aims to find scenes with the same content across heterogeneous sensors, such as SAR and optical or panchromatic (PAN) and multispectral (MS) imagery, without manual labels. Existing self-supervised methods either reconstruct sensor-specific pixel detail, \TANI{align each image only with its co-registered twin}, or rely on large transformers with 118--173M parameters (and up to 398M). We introduce RS-JEPA, a compact joint-embedding predictive framework built on a Vision Mamba encoder shared by both sensors. Since a state-space encoder reads an image as an ordered sequence of patches, we propose scan-path views, which treat two reading orders of the same image as two views with identical pixels and predict the hidden patches in one order from the visible patches in the other.To keep this prediction challenging, we propose Adversarial View Coupling (AVC), which learns the hardest split of the visible patches, and a neighbourhood consensus loss that aligns the scene similarities of both sensors. Extensive experiments on SAR--optical (BEN-14K, CBRSIR-VS) and PAN--MS (DSRSID) benchmarks demonstrate that RS-JEPA achieves the highest F1@5 among twelve re-implemented methods (72.92, 80.35 and 94.43) with a retrieval network of only 3.88M parameters (14.57M in training, \TANI{8--16$\times$} fewer than prior JEPA retrievers). \TANI{Like prior methods, RS-JEPA trains on co-registered pairs.} Our results highlight reading order as a source of self-supervised views. code~\url{https://github.com/Tanishq251/RSJEPA}.

%% file: sec/1_introduction.tex
\section{Introduction}
\label{sec:introduction}

\begin{figure}[t]
\label{fig:efficiency}\centering\includegraphics[width=0.48\textwidth]{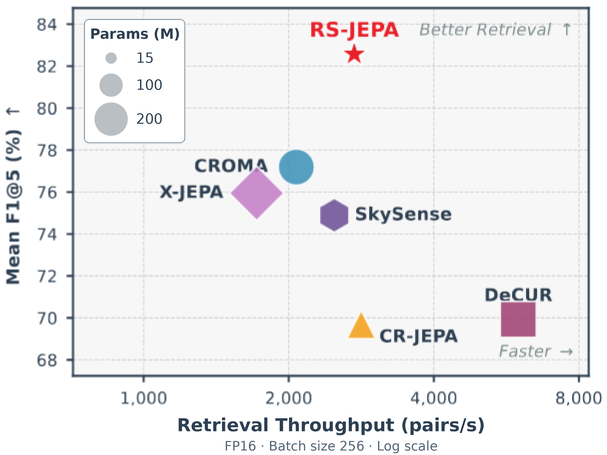}
  \caption{\textbf{Accuracy against model size.} Mean F1@5 over the three benchmarks \TANI{against retrieval throughput (marker area: parameters).}}
  \vspace{-5mm}
\end{figure}

Synthetic aperture radar (SAR) images the ground through cloud and at night, whereas optical sensors provide detailed multispectral or panchromatic images only under clear skies~\cite{torres2012sentinel1,drusch2012sentinel2,li2018dsrsid}.
\emph{Cross-sensor image retrieval} ranks the images of one sensor by their content similarity to a query from another~\cite{sumbul2021bigearthnetmm}.
This is difficult because the same scene looks very different across sensors, labels are scarce at archive scale, and every archive image must be embedded, which favours small encoders.
Masked image models~\cite{he2022mae,cong2022satmae,noman2024satmaepp} reconstruct pixels and thus also learn sensor noise such as SAR speckle.
Cross-sensor alignment models~\cite{fuller2023croma,wang2024decur} match each image with its co-registered counterpart but ignore how different scenes relate.
Joint-embedding predictive architectures (JEPAs)~\cite{assran2023ijepa} predict features instead of pixels, and X-JEPA~\cite{xjepa} and CR-JEPA~\cite{crjepa} lead label-free retrieval on BEN-14K this way, but rely on large transformers that are costly to run over whole archives.
State-space models (SSMs)~\cite{gu2023mamba,zhu2024vim} are far smaller, yet they read an image as an ordered sequence, while existing self-supervised views were designed for order-agnostic attention.

\begin{tcolorbox}[
colback=yellow!8, colframe=yellow!50!black,
boxrule=0.8pt, arc=3mm, left=1mm,
right=1mm, top=1mm, bottom=1mm
]
\centering
\textit{\textbf{{\color{orange!85!yellow}\faLightbulb}
Can we design a compact cross-sensor retrieval model that learns what a scene contains, regardless of the sensor, simply by reading it in two different orders?
}}
\end{tcolorbox}

A selective SSM summarises only the tokens it has already read, so its output for a patch depends on how the patch grid is flattened~\cite{gu2023mamba,zhu2024vim}.
Two traversals of one image therefore give two feature sequences from identical pixels, which is a natural pair of views.
Only scene content stays predictable across paths; speckle and other sensor texture do not.
Since prediction happens in feature space and the views come from the encoder itself, no image editing or momentum encoder is needed.

Existing vision SSMs fix the traversal or combine several directions~\cite{liu2024vmamba,ren2025arm}, discarding reading order before it can supervise learning.
Self-supervised remote sensing methods build views by masking, augmentation or sensor pairing~\cite{assran2023ijepa,fuller2023croma,wang2024decur}, assuming order-insensitive encoders.
To our knowledge, no prior work has shown whether reading order can generate views for cross-sensor joint-embedding prediction, or how to train such views for retrieval.

Two complementary challenges must be addressed together.
\textbf{First}, two scan paths over the same visible patches share much local context, so the predictor can exploit spatial continuity instead of modelling the scene, making the task too easy.
\textbf{Second}, instance-level contrastive alignment pulls each scene towards its twin but leaves other scenes unconstrained, so the two sensors may induce different neighbourhoods, which directly harms retrieval.

To address these challenges, we introduce RS-JEPA, a remote sensing JEPA that learns cross-sensor features from the reading order of one Vision Mamba encoder shared by both sensors.
Its scan-path views encode the visible patches along one traversal and predict the stop-gradient features of the hidden patches along another.
For the first challenge, Adversarial View Coupling (AVC) learns, through gradient reversal, the split of visible patches that is hardest to predict and couples the two halves with cross-view attention.
For the second, a neighbourhood consensus loss aligns the same- and cross-sensor similarity distributions of both sensors.
RS-JEPA achieves the highest mean F1@5 among twelve re-implemented methods on three benchmarks with a retrieval network of only 3.88M parameters.

\noindent Our key contributions are as follows:
\begin{itemize}\setlength{\itemsep}{0pt}\setlength{\parsep}{0pt}
  \item We propose \textbf{RS-JEPA}, a compact joint-embedding predictive framework for label-free cross-sensor retrieval with one shared Vision Mamba encoder.
  \item We introduce \textbf{scan-path views}, which predict the hidden patches in one order from the visible patches in another, without image editing or a momentum encoder.
  \item We design \textbf{Adversarial View Coupling} (AVC), which learns the hardest split of the visible patches into two views, and a teacher-free \textbf{neighbourhood consensus} loss that aligns the scene similarities of both sensors.
  \item Extensive experiments on \textbf{BEN-14K, CBRSIR-VS and DSRSID} demonstrate state-of-the-art retrieval performance with a substantially smaller model.
\end{itemize}

%% file: sec/2_related_works.tex
\section{Related Work}
\label{sec:related}

\textbf{Self-Supervised Learning in Remote Sensing:}
Masked image modelling~\cite{he2022mae} has been adapted to satellite imagery with spectral and temporal grouping~\cite{cong2022satmae}, multiple scales~\cite{noman2024satmaepp,reed2023scalemae,tang2023crossscalemae} and rotated windows~\cite{wang2023maervsa}.
For radar--optical pairs, CROMA~\cite{fuller2023croma} combines redrawing with cross-sensor matching, DeCUR~\cite{wang2024decur} separates shared from sensor-specific features, and others use self-distillation~\cite{wang2022dinomm} or fusion~\cite{chen2022sarfusion}; SkySense~\cite{guo2024skysense} and RoMA~\cite{wang2025roma} scale this to foundation models.
They target fine-tuning for classification or segmentation, not small retrieval encoders.

\textbf{Cross-Sensor Retrieval:}
Early methods learned binary hash codes for radar and optical images from labels~\cite{xiong2020cmhash,cbrsirvs} on paired benchmarks such as BigEarthNet-MM~\cite{sumbul2019bigearthnet,sumbul2021bigearthnetmm}.
CSMAE~\cite{hackstein2024csmae} removed the labels with cross-sensor masked autoencoders, and X-JEPA~\cite{xjepa} and CR-JEPA~\cite{crjepa} replaced pixel redrawing with cross-sensor feature prediction in large transformers; RS-JEPA uses a compact state-space encoder. While REJEPA~\cite{choudhury2025rejepa} predicts masked patch embeddings within a single image for same-sensor retrieval, RS-JEPA uses adversarially split scan views and symmetric neighbourhood consensus to align scene similarities across radar and optical embeddings.

\textbf{Views and Relations in Self-Supervised Learning:}
Rather than relying on heuristic image augmentations or asymmetric teacher-student models~\cite{assran2023ijepa,ferreira2025hvs}, RS-JEPA couples adversarial view splitting (AVC) with symmetric neighbourhood consensus to align cross-sensor similarity geometry without external guidance.

\textbf{State-Space Models for Vision:}
Vision Mamba (Vim)~\cite{zhu2024vim} adapts selective state-space models~\cite{gu2023mamba} to vision by traversing patch sequences bidirectionally, while VMamba~\cite{liu2024vmamba} incorporates four-way scanning directions, and autoregressive models~\cite{ren2025arm} enforce sequential ordering.
In prior works, the scan trajectory is treated as a fixed architectural choice; RS-JEPA instead leverages multi-directional scan paths as an intrinsic source of multi-view self-supervision.

%% file: sec/3_methodology.tex
\section{Proposed Method}
\label{sec:methodology}

In this section, we present RS-JEPA, which learns cross-sensor retrieval features without labels by predicting the features of hidden patches, rather than their pixels, with a single Vision Mamba encoder shared by both sensors.

\subsection{Overview}
\label{sec:method_problem}

\begin{figure*}[t]
  \centering\includegraphics[width=0.9\textwidth]{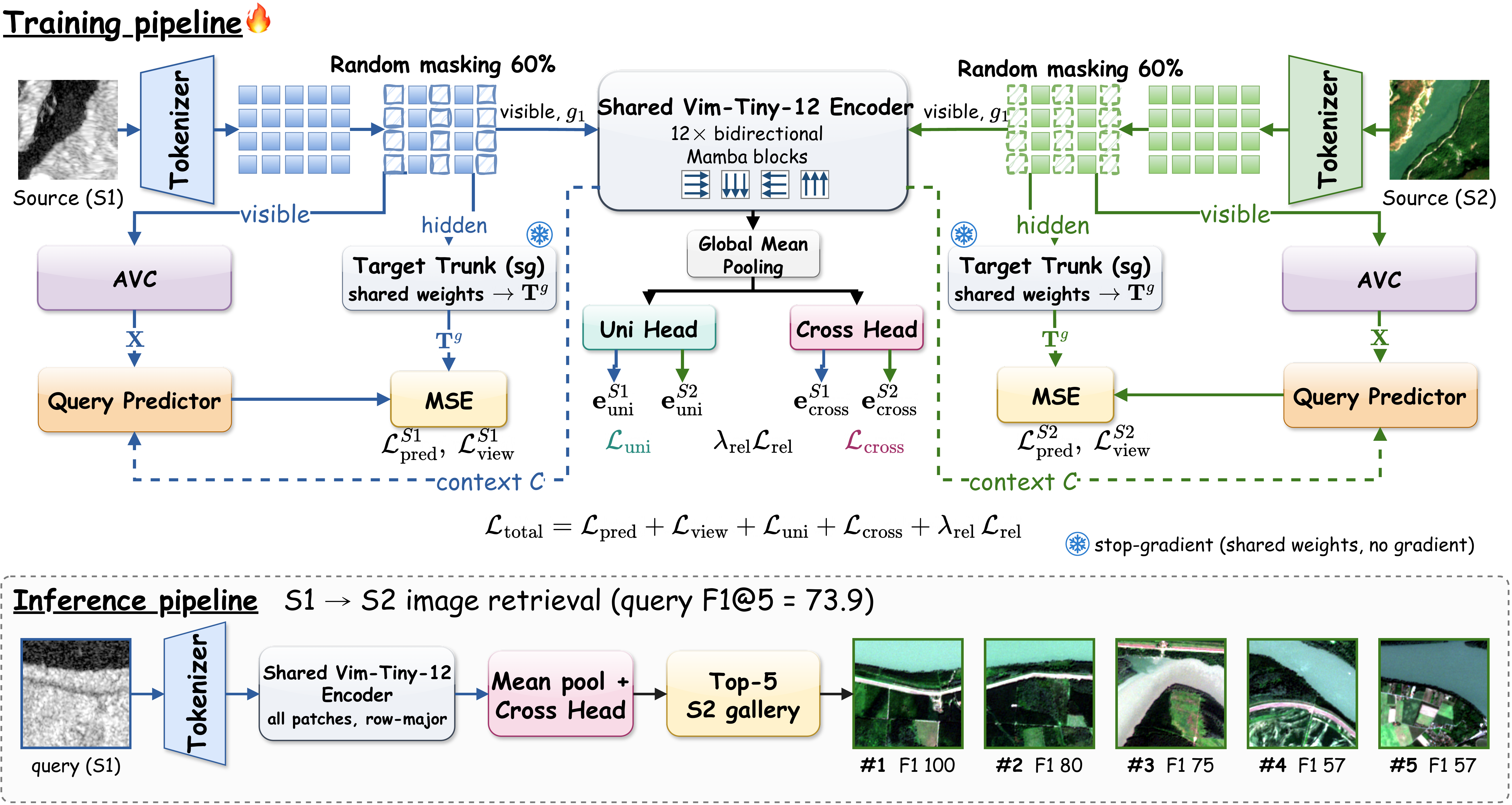}
  \caption{\textbf{Overview of RS-JEPA.} \emph{Top:} training. Per-sensor stems feed one shared Vim encoder; AVC (\cref{fig:avc}) and a path-conditioned predictor (\cref{fig:blocks}b) predict stop-gradient targets $\mathbf{T}^{g}$, and the heads are aligned by InfoNCE and neighbourhood consensus\ADD{ (full objective in \cref{eq:total})}. \emph{Bottom:} inference, a real S1$\rightarrow$S2 top-5 on BEN-14K test (query with F1@5 closest to the mean, not hand-picked).}
  \label{fig:arch}
\end{figure*}

Given $M$ unlabeled, co-registered pairs $\{(x^{(1)}_i, x^{(2)}_i)\}_{i=1}^{M}$ from sensors S1 (\eg, SAR or panchromatic) and S2 (\eg, optical or multispectral), indexed by $m\in\{1,2\}$, we seek an encoder whose unit-norm embeddings rank semantically related scenes first under cosine similarity in the same-sensor (S1$\to$S1, S2$\to$S2) and cross-sensor (S1$\to$S2, S2$\to$S1) directions.
A sensor-specific stem (strided convolution plus learned position embeddings $\mathbf{p}_j$) maps each image \ADD{$x^{(m)}$} to $N$ patch tokens $\mathbf{Z}^{(m)}\in\mathbb{R}^{N\times D}$, and a random mask splits them into a visible set $\mathcal{V}$ and a hidden set $\mathcal{H}$ with $|\mathcal{H}|\approx rN$\ADD{ for masking ratio $r$}, giving $\mathbf{Z}_{\mathcal{V}}$ and $\mathbf{Z}_{\mathcal{H}}$.
As shown in \cref{fig:arch}, a shared 12-block Vision Mamba (Vim) encoder $f_\theta$~\cite{zhu2024vim} of width \ADD{$D$} reads the visible tokens along one scan path, and a per-sensor predictor predicts the hidden tokens as the same encoder reads them along \ADD{that path and along another} (\cref{sec:method_scanpath}).
AVC makes this prediction harder (\cref{sec:method_avc}), and retrieval heads on the pooled features align the two sensors (\cref{sec:method_consensus,sec:method_objective}).
The encoder and heads are shared by both sensors, whereas the stems, AVC modules and predictors are per sensor; $\mathrm{sg}[\cdot]$ denotes stop-gradient.

\subsection{Scan-Path Views}
\label{sec:method_scanpath}

\begin{figure}[h]
  \centering\includegraphics[width=0.86\columnwidth]{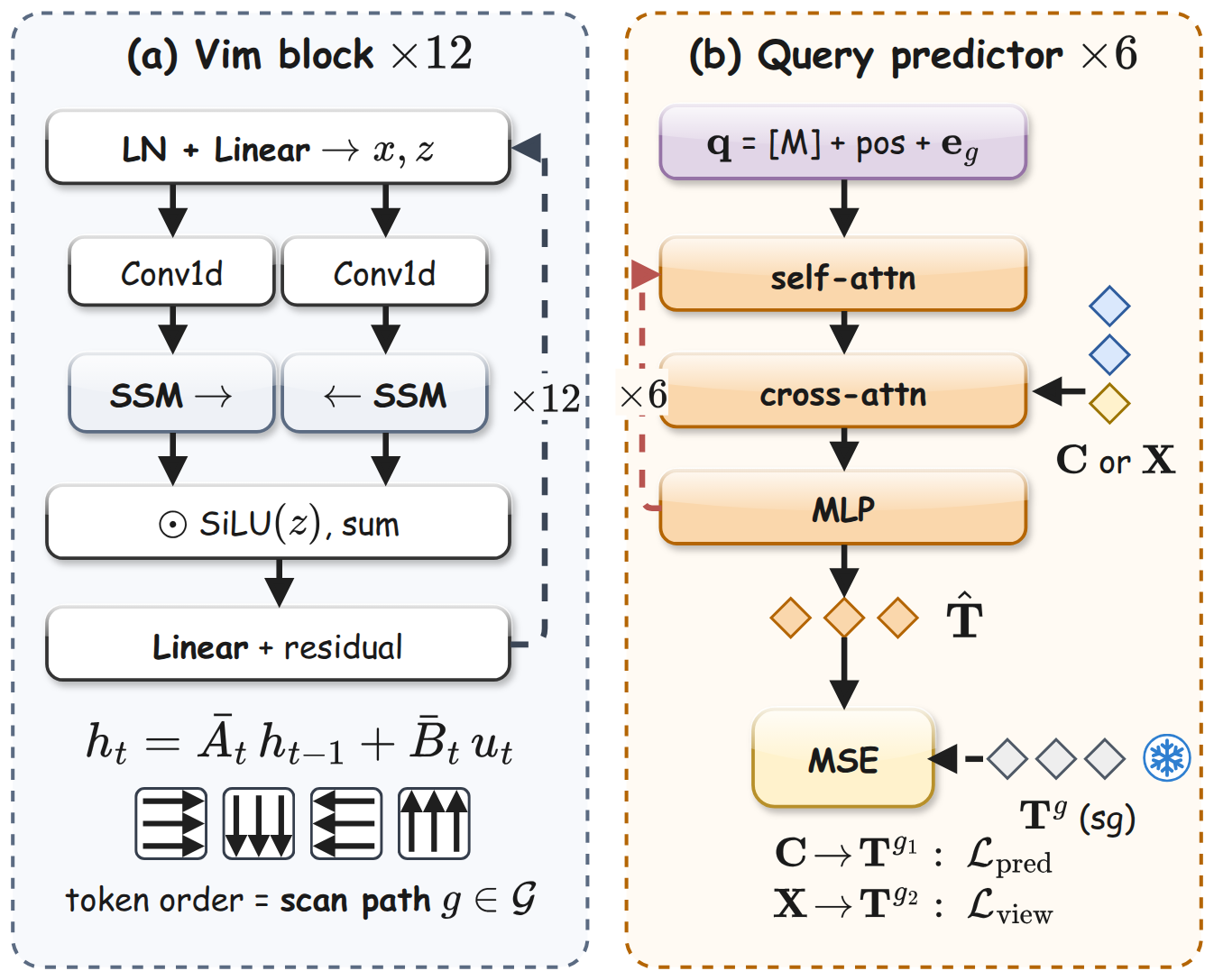}
  \caption{\textbf{Encoder and predictor.} (a)~One of the 12 Vim blocks: bidirectional selective scans along path $g$. (b)~The six-layer query predictor; queries carry position and path code $\mathbf{e}_g$.}
  \label{fig:blocks}
\end{figure}

Since a state-space encoder reads an image as an ordered sequence, two traversals of the same image yield two views with identical pixels, which we use \ADD{as self-supervised views without image editing}.
\vspace{-3mm}

\paragraph{Order dependence of selective scans.}
A selective state-space layer processes a token sequence $(\ADD{s}_1,\dots,\ADD{s}_L)$ with the input-dependent recurrence
\vspace{-1 pt}
\begin{equation}
  h_t=\bar{A}_t\,h_{t-1}+\bar{B}_t\,\ADD{s}_t,\qquad y_t=\ADD{\Gamma_t}\,h_t,
  \label{eq:ssm}
\end{equation}
where $h_t$ is the hidden state, and the discretised $\bar{A}_t,\bar{B}_t$ and the readout $\ADD{\Gamma_t}$ are functions of $\ADD{s}_t$. Because $h_t$ summarises exactly the tokens that precede $t$, the output for a patch depends on the order in which the 2-D grid is flattened, and running the recurrence in both directions, as Vim does, still depends on the chosen traversal.

\vspace{-0.4 cm}

\paragraph{Path-ordered encoding (\cref{fig:blocks}a).}
Let $\mathcal{G}$ be a set of four traversals of the patch grid: row-major, column-major, and a mirrored version of each (row-major mirrored left to right, column-major mirrored top to bottom). Because Vim also reads each traversal in reverse, they give four distinct scan dynamics. Each $g\in\mathcal{G}$ assigns every patch $j$ a rank $\rho_g(j)$. Encoding a token subset along $g$, written $f_\theta(\cdot\,;g)$, sorts the tokens by $\rho_g$ of their patch indices, applies the trunk, and restores the original order. Position embeddings are added before sorting, so only the scan order changes. Per image, the context path $g_1$ is drawn uniformly from $\mathcal{G}$, and
\begin{equation}
  \mathbf{C}=f_\theta(\mathbf{Z}_{\mathcal{V}};g_1),\qquad
  \mathbf{T}^{g}=\mathrm{sg}\big[f_\theta(\mathbf{Z}_{\mathcal{H}};g)\big]\ \ \forall g\in\mathcal{G}.
  \label{eq:ctx_tgt}
\end{equation}
The targets are the hidden tokens encoded on their own by the \emph{same} trunk, without a momentum encoder; this costs one gradient-free trunk pass over the hidden tokens per path during training and nothing at inference. The second path is the one whose targets disagree most with those of $g_1$:
\begin{equation}
  g_2=\operatorname*{arg\,max}_{g\in\mathcal{G}\setminus\{g_1\}}\ \tfrac{1}{|\mathcal{H}|D}\,\big\|\mathbf{T}^{g}-\mathbf{T}^{g_1}\big\|_F^2 .
  \label{eq:g2}
\end{equation}
In an earlier configuration (50\% masking, without AVC), a random $g_2\neq g_1$ performed comparably on BEN-14K (72.63 \vs 72.56 F1@5), so we do not count this rule as a contribution.
\vspace{-0.5 cm}

\begin{figure*}[!h]
  \centering\includegraphics[width=0.7\textwidth, height=0.3\textwidth]{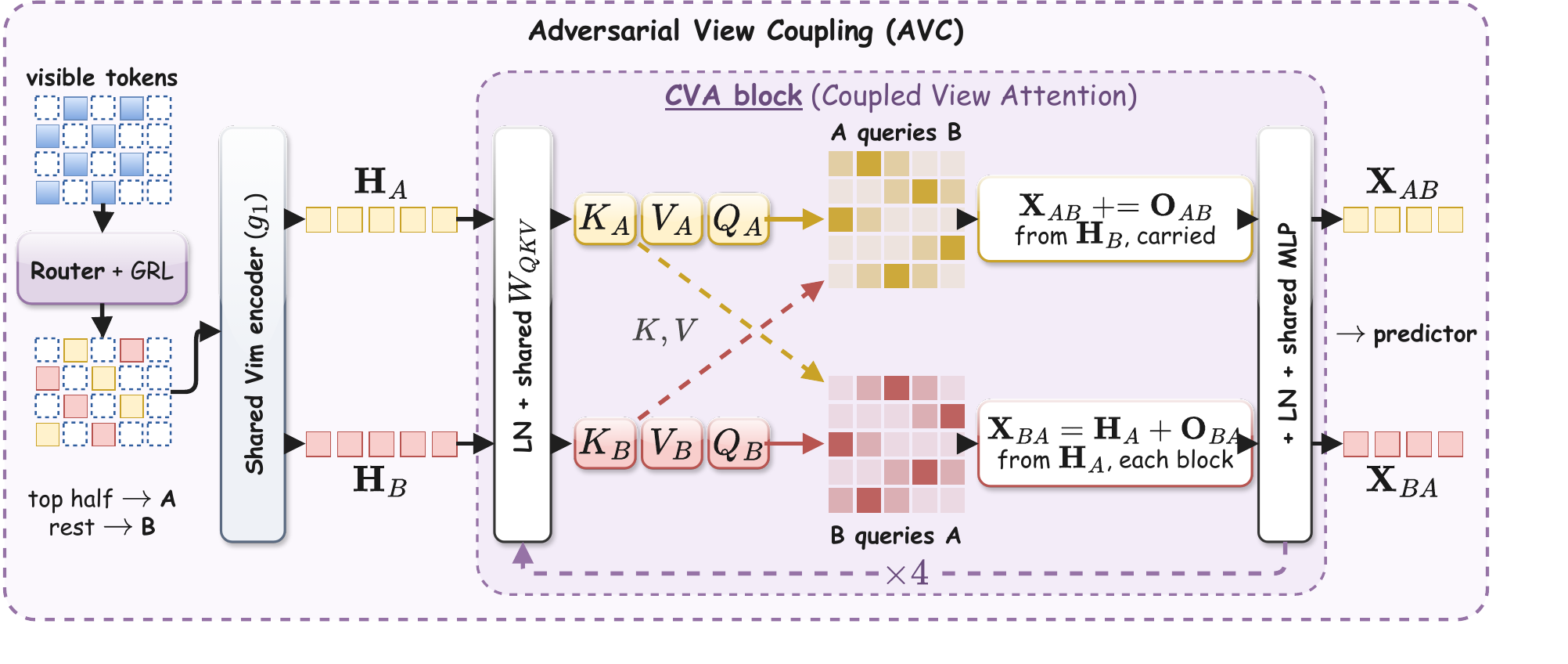}
  \vspace{-3mm}
  \caption{\textbf{Adversarial View Coupling.} A scorer with gradient reversal splits the visible tokens into views $A$ and $B$, which the shared encoder reads along $g_1$. In each of four CVA blocks, one view's queries attend to the other view's keys and values, giving $\mathbf{X}_{AB}$ and $\mathbf{X}_{BA}$ for the predictor.}
  \label{fig:avc}
\end{figure*}

\paragraph{Path-conditioned prediction (\cref{fig:blocks}b).}
The predictor $P^{(m)}$ of sensor $m$ is a transformer\ADD{~\cite{vaswani2017attention}} whose queries for the hidden patches are $\mathbf{q}_j=\mathbf{z}_{\mathrm{mask}}+\mathbf{p}_j+\mathbf{e}_g$ for $j\in\mathcal{H}$, where $\mathbf{z}_{\mathrm{mask}}$ is a learned mask token and $\mathbf{e}_g\in\mathbb{R}^D$ is a learned path code that tells the predictor which traversal to predict. Each layer applies self-attention among the queries, cross-attention to a conditioning sequence and an MLP. With $P^{(m)}(\mathbf{S};g)$ denoting the prediction from conditioning $\mathbf{S}$ for path $g$, the context loss is
\begin{equation}
  \mathcal{L}_{\mathrm{ctx}}^{(m)}=\tfrac{1}{|\mathcal{H}|D}\,\big\|P^{(m)}(\mathbf{C};g_1)-\mathbf{T}^{g_1}\big\|_F^2 .
  \label{eq:lctx}
\end{equation}
The view loss $\mathcal{L}_{\mathrm{view}}^{(m)}$ (\cref{eq:lview}) instead predicts the $g_2$ targets from the AVC streams, so only information that survives a change of path, namely the content of the scene, can be predicted reliably.

\subsection{Adversarial View Coupling}
\label{sec:method_avc}

Two scan paths over the same visible patches share much local context, so their prediction can become trivial; AVC prevents this by learning the split of the visible patches \ADD{from whose two halves the hidden-patch targets are hardest to predict}.
\vspace{-0.4 cm}

\paragraph{Adversarial view partition (\cref{fig:avc}).}
A per-sensor linear scorer gives every visible stem token a score $a_j=\mathbf{w}^{\top}\mathbf{z}_j+b$, $j\in\mathcal{V}$. The top-scoring half of the visible tokens forms view $\mathcal{V}_A$ and the rest form $\mathcal{V}_B$, a hard, disjoint split of equal size. As the split itself is not differentiable, the scorer learns through a straight-through weight:
\begin{equation}
\footnotesize
  \tilde{\mathbf{z}}_j=\mathbf{z}_j\,\big(1+\omega_j-\mathrm{sg}[\omega_j]\big),\quad
  \omega_j=\begin{cases}\sigma(\mathcal{R}(a_j)), & j\in\mathcal{V}_A,\\ 1-\sigma(\mathcal{R}(a_j)), & j\in\mathcal{V}_B,\end{cases}
  \label{eq:route}
\end{equation}
whose forward value is exactly one. Here $\sigma$ is the logistic sigmoid and $\mathcal{R}$ is a gradient-reversal layer, which acts as the identity in the forward pass and negates the gradient in the backward pass. The predictor and encoder minimise $\mathcal{L}_{\mathrm{view}}$, while the scorer\ADD{ receives the negated gradient and passes it to} the stem tokens it reads\ADD{, in addition to their ordinary gradient}, so the scorer learns the partition that makes this prediction hardest.
\vspace{-0.3 cm}

\paragraph{Shared encoding and Coupled View Attention.}
Both views pass through the \emph{same} trunk along the context path, $\mathbf{H}_A=f_\theta(\tilde{\mathbf{Z}}_{\mathcal{V}_A};g_1)$ and $\mathbf{H}_B=f_\theta(\tilde{\mathbf{Z}}_{\mathcal{V}_B};g_1)$, so the coupling gradient trains the retrieval backbone itself. The two views are then coupled by a stack of \ADD{four} Coupled View Attention (CVA) blocks (\cref{fig:avc}) whose projections and MLP are shared by both streams. The streams $\mathbf{H}_A,\mathbf{H}_B$ are refined by self-attention, and their queries, keys and values $(\mathbf{Q}_v,\mathbf{K}_v,\mathbf{V}_v)$, $v\in\{A,B\}$, enter a single coupled attention call
\begin{equation}
  \begin{bmatrix}\mathbf{O}_{AB}\\ \mathbf{O}_{BA}\end{bmatrix}
  =\mathrm{Attn}\!\left(\begin{bmatrix}\mathbf{Q}_A\\ \mathbf{Q}_B\end{bmatrix},
  \begin{bmatrix}\mathbf{K}_B\\ \mathbf{K}_A\end{bmatrix},
  \begin{bmatrix}\mathbf{V}_B\\ \mathbf{V}_A\end{bmatrix}\right),
  \label{eq:cva}
\end{equation}
where rows are stacked along the batch axis, so view-$A$ queries attend only to view-$B$ keys and values, and vice versa. Each output is added residually and passed through a shared MLP, giving the crossed streams $\mathbf{X}_{AB}$ (initialised from $\mathbf{H}_B$ and carried across blocks) and $\mathbf{X}_{BA}$ (recomputed from $\mathbf{H}_A$ in every block)\ADD{, so only $\mathbf{X}_{AB}$ accumulates context over depth}. The predictor then predicts the targets of the \emph{other} path from the crossed streams $[\mathbf{X}_{AB};\mathbf{X}_{BA}]$\ADD{, concatenated along the token axis}:
\begin{equation}
  \mathcal{L}_{\mathrm{view}}^{(m)}=\tfrac{1}{|\mathcal{H}|D}\,\big\|P^{(m)}([\mathbf{X}_{AB};\mathbf{X}_{BA}];g_2)-\mathbf{T}^{g_2}\big\|_F^2 .
  \label{eq:lview}
\end{equation}
The encoder must thus produce features that remain predictive across adversarially chosen subsets of the scene.

\subsection{Cross-Sensor Neighbourhood Consensus}
\label{sec:method_consensus}

Instance-level contrastive alignment pulls each scene towards its co-registered twin but leaves the arrangement of the other scenes unconstrained; the consensus loss instead requires both sensors to agree on which scenes are neighbours.
For a batch of $\ADD{K}$ pairs, let $\mathbf{u}^{(m)}_i$ and $\mathbf{c}^{(m)}_i$ be the unit-norm \emph{uni}- and \emph{cross}-head embeddings of scene $i$ from sensor $m$ (\cref{sec:method_objective}). The same-sensor neighbourhood distribution of scene $i$ over the other scenes is
\begin{equation}
  \ADD{Q}_m(j\,|\,i)=\frac{\exp(\mathbf{u}^{(m)\top}_i\mathbf{u}^{(m)}_j/\tau_r)}{\sum_{k\neq i}\exp(\mathbf{u}^{(m)\top}_i\mathbf{u}^{(m)}_k/\tau_r)},\quad j\neq i,
  \label{eq:nbr}
\end{equation}
where $\tau_r$ is a relational temperature. The cross-sensor distribution $\ADD{Q}_{m\to m'}(j\,|\,i)$ is defined in the same way, with query $\mathbf{c}^{(m)}_i$ and gallery $\{\mathbf{c}^{(m')}_j\}_{j\neq i}$ for $m'\neq m$. The target is the consensus of the two sensors, $\ADD{\bar{Q}_i=}\mathrm{sg}\big[\tfrac12\sum_{m}\ADD{Q}_m(\cdot\,|\,i)\big]$, and all four distributions are pulled towards it:
\begin{equation}
\begin{aligned}
  \mathcal{L}_{\mathrm{rel}}=\frac{1}{4\ADD{K}}\sum_{i=1}^{\ADD{K}}\Big[&\sum_{m}\mathrm{KL}\big(\ADD{\bar{Q}}_i\,\|\,\ADD{Q}_m(\cdot|i)\big)\\
  +&\sum_{m\neq m'}\mathrm{KL}\big(\ADD{\bar{Q}}_i\,\|\,\ADD{Q}_{m\to m'}(\cdot|i)\big)\Big].
\end{aligned}
  \label{eq:lrel}
\end{equation}
The twin $j=i$ is excluded from every distribution, and the loss vanishes \ADD{only when all four distributions coincide}.

\subsection{Training Objective and Inference}
\label{sec:method_objective}

\vspace{-2mm}
\paragraph{Retrieval heads and inference.}
The context $\mathbf{C}^{(m)}$ is mean-pooled and passed through two MLP heads shared by both sensors, \emph{uni} and \emph{cross}, whose $\ell_2$-normalised outputs are $\mathbf{u}^{(m)}_i$ and $\mathbf{c}^{(m)}_i$. Both heads are trained with the symmetric InfoNCE loss\ADD{~\cite{oord2018representation}} $\mathcal{L}_{\mathrm{nce}}$ with temperature $\tau$ and in-batch negatives, and the uni head also with a cosine term $\mathcal{L}_{\cos}=1-\tfrac{1}{\ADD{K}}\sum_i\mathbf{u}^{(1)\top}_i\mathbf{u}^{(2)}_i$. At test time, the encoder reads all $N$ patches along the row-major path without masking, and gallery items are ranked by cosine similarity of the uni (same-sensor) or cross (cross-sensor) embeddings. The deployed model consists only of the two stems, the shared trunk and the two heads.

\vspace{-1mm}

\paragraph{Full objective.}
The total loss is
\begin{equation}
\begin{aligned}
  \mathcal{L}={}&\textstyle\sum_{m}\mathcal{L}_{\mathrm{ctx}}^{(m)}+\tfrac12\sum_{m}\mathcal{L}_{\mathrm{view}}^{(m)}
  +\mathcal{L}_{\mathrm{nce}}(\mathbf{c}^{(1)},\mathbf{c}^{(2)})\\
  &+\mathcal{L}_{\mathrm{nce}}(\mathbf{u}^{(1)},\mathbf{u}^{(2)})+\mathcal{L}_{\cos}+\lambda_{\mathrm{rel}}\,\mathcal{L}_{\mathrm{rel}},
\end{aligned}
  \label{eq:total}
\end{equation}
where each loss term is averaged over the batch and $\lambda_{\mathrm{rel}}$ weights the consensus term. All parameters, including the scorers through the reversed gradient of \cref{eq:route}, are optimised jointly in a single stage.\ADD{ The in-batch negatives of $\mathcal{L}_{\mathrm{nce}}$ prevent collapse.}

%% file: sec/4_experiments.tex
\section{Experiments}
\label{sec:experiments}

\input{sec/5_auto_qualcap}
\begin{table*}[t]
\centering
\small
\setlength{\tabcolsep}{2.3pt}
\caption{\textbf{Cross-sensor retrieval on BEN-14K, CBRSIR-VS and DSRSID.} F1@5 (\%); $i{\to}j$: query from sensor S$i$, gallery from sensor S$j$. Best in \textbf{bold}, second best \underline{underlined}. All methods are re-implemented and trained by us under one protocol. Params: training model; GFLOPs: one S1--S2 pair at $120\times120$ (2 FLOPs per multiply-accumulate; selective scan included for RS-JEPA), BEN-14K configuration.}
\label{tab:main}
\resizebox{\textwidth}{!}{%
\input{tables/table1_main}

}
\par\vspace{1.2em}
\centering\includegraphics[width=0.9\textwidth]{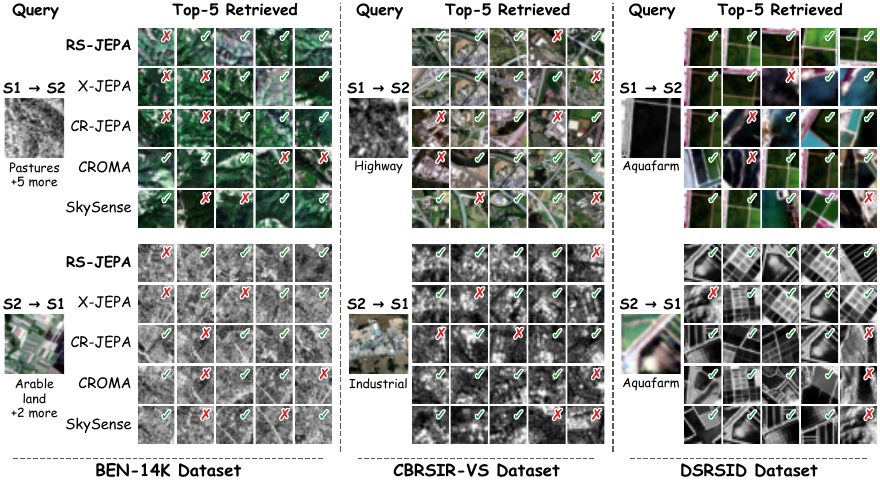}\par
\vspace{-0.9em}
\captionof{figure}{\qualcap}
\label{fig:qualitative}
\end{table*}

\input{sec/4_1_datasets}

\subsection{Implementation Details}
\label{sec:setup}\label{sec:training}
RS-JEPA uses a 12-block Vim encoder (width 192) on $10\times10$ patches, a fixed 60\% masking ratio~\cite{assran2023ijepa,xjepa}, a six-layer predictor and four CVA blocks, with $\tau=0.07$, $\tau_r=0.1$ and $\lambda_{\mathrm{rel}}=1$. We train for 100 epochs with AdamW (batch 290, learning rate warmed up to $10^{-3}$ over 15 epochs, then cosine decay) and identical random flips and $90^\circ$ rotations on both sensors, and report results across independent runs. All experiments run in PyTorch on four NVIDIA V100 GPUs with 32 GB RAM and an Intel Xeon CPU.

\noindent\textbf{Baselines:} DeCUR uses a ResNet-50 backbone, as in its original design~\cite{wang2024decur}, and the other seven use a ViT-Base/16 backbone; all eight are re-implemented following the X-JEPA protocol~\cite{xjepa} and trained from scratch by us for 100 epochs on the same pairs at $120\times120$, with shared retrieval code; single-sensor MAE variants get one encoder per sensor and an InfoNCE head. \cref{tab:main} thus compares methods under one training budget, not their best published results (\eg, CR-JEPA reports 75.82/75.40 cross-sensor F1@5 on BEN-14K after 400 epochs at $224\times224$~\cite{crjepa}); at their original $224\times224$ setting, X-JEPA and CR-JEPA still trail RS-JEPA (\cref{fig:resolution}c).

\subsection{Evaluation Protocol}
\label{sec:protocol}
Following X-JEPA~\cite{xjepa}, we evaluate four retrieval directions (S1$\to$S1, S2$\to$S2, S1$\to$S2, S2$\to$S1) and their mean. The main metric is F1@5 between the label sets of the query and its top-5 results, which equals precision@5 on the single-label CBRSIR-VS and DSRSID.
For mAP, a result is relevant if it shares a label with the query.
Same-sensor search excludes the query, while the cross-sensor gallery keeps its co-registered partner~\cite{xjepa,crjepa}; \cref{sec:results} also reports results without it.

%% file: sec/5_auto_qualcap.tex
\newcommand{\qualcap}{\textbf{Qualitative retrieval.} Comparison of the top-5 results of each method for one S1$\rightarrow$S2 and one S2$\rightarrow$S1 query per benchmark. \cmark: relevant (F1 $\geq 0.5$ on BEN-14K; same class otherwise), \xmark: irrelevant. Queries are selected by a fixed rule as the most representative of each method's F1@5, not hand-picked.}


%% file: tables/table1_main.tex
\begin{tabular}{@{}lrr>{\hspace{6pt}}ccccc>{\hspace{6pt}}ccccc>{\hspace{6pt}}ccccc@{}}
\toprule
\multirow{2}{*}{Method} & \multicolumn{1}{c}{Params} & \multicolumn{1}{c}{GFLOPs} & \multicolumn{5}{c}{BEN-14K} & \multicolumn{5}{c}{CBRSIR-VS} & \multicolumn{5}{c}{DSRSID} \\
\cmidrule(lr){4-8} \cmidrule(lr){9-13} \cmidrule(l){14-18}
& \multicolumn{1}{c}{(M) $\downarrow$} & \multicolumn{1}{c}{$\downarrow$} & 1$\to$1 & 2$\to$2 & 1$\to$2 & 2$\to$1 & Avg. & 1$\to$1 & 2$\to$2 & 1$\to$2 & 2$\to$1 & Avg. & 1$\to$1 & 2$\to$2 & 1$\to$2 & 2$\to$1 & Avg. \\
\midrule
\multicolumn{18}{@{}l}{\textit{Masked image modelling}} \\
MAE-RVSA\cite{wang2023maervsa} & 172.23 & 50.33 & 65.27 & 68.87 & 68.28 & 67.88 & 67.58 & 64.28 & 73.35 & 63.41 & 64.79 & 66.46 & 74.94 & 73.10 & 77.86 & 77.45 & 75.84 \\
ScaleMAE\cite{reed2023scalemae} & 215.07 & 51.12 & 63.88 & 68.93 & 67.86 & 68.93 & 67.40 & 64.99 & 75.29 & 64.59 & 66.10 & 67.74 & 73.47 & 71.84 & 76.12 & 75.49 & 74.23 \\
SatMAE++\cite{noman2024satmaepp} & 223.57 & 106.18 & 64.85 & 69.08 & 67.69 & 68.27 & 67.47 & 64.93 & 74.55 & 65.47 & 65.23 & 67.55 & 77.07 & 77.00 & 78.29 & 78.80 & 77.79 \\
\midrule
\multicolumn{18}{@{}l}{\textit{Multi-sensor self-supervised}} \\
CROMA\cite{fuller2023croma} & 200.27 & 38.17 & \underline{68.57} & 71.09 & 68.92 & 69.97 & 69.64 & 72.16 & 80.13 & 72.08 & 73.65 & 74.50 & 87.53 & \underline{88.14} & 87.28 & 86.74 & 87.42 \\
DeCUR\cite{wang2024decur} & 198.07 & 5.22 & 66.06 & 66.43 & 62.39 & 61.48 & 64.09 & \underline{75.99} & \underline{85.92} & \underline{72.83} & \underline{77.55} & \underline{78.07} & 73.67 & 73.31 & 62.49 & 61.05 & 67.63 \\
SkySense\cite{guo2024skysense} & 173.24 & 51.13 & 66.65 & 71.01 & 67.97 & 68.02 & 68.41 & 66.23 & 74.28 & 65.70 & 65.04 & 67.81 & \underline{87.94} & 88.02 & \underline{88.96} & \underline{88.85} & \underline{88.44} \\
\midrule
\multicolumn{18}{@{}l}{\textit{Joint-embedding predictive}} \\
CR-JEPA\cite{crjepa} & 115.18 & 14.74 & 67.93 & \underline{71.64} & \underline{69.27} & \underline{70.41} & \underline{69.81} & 60.77 & 74.72 & 61.64 & 64.73 & 65.46 & 71.55 & 73.65 & 74.70 & 75.25 & 73.79 \\
X-JEPA\cite{xjepa} & 230.47 & 32.98 & 66.27 & 70.08 & 66.81 & 68.01 & 67.79 & 67.94 & 81.50 & 71.78 & 71.52 & 73.18 & 85.84 & 85.91 & 87.81 & 87.89 & 86.86 \\
\midrule
\rowcolor{oursbg}\textbf{RS-JEPA (Ours)} & 14.57 & 2.23 & \textbf{70.93} & \textbf{72.55} & \textbf{73.86} & \textbf{74.33} & \textbf{72.92} & \textbf{76.39} & \textbf{86.54} & \textbf{78.72} & \textbf{79.75} & \textbf{80.35} & \textbf{93.77} & \textbf{93.73} & \textbf{95.10} & \textbf{95.14} & \textbf{94.43} \\
\bottomrule
\end{tabular}

%% file: sec/4_1_datasets.tex
\subsection{Datasets and Data Splits}
\label{sec:datasets}
We use three dual-sensor benchmarks; training is label-free and train and test scenes do not overlap.
\textbf{BEN-14K}~\cite{sumbul2019bigearthnet}, a BigEarthNet subset over ten European countries, has 14{,}832 Sentinel-1 (VV/VH, S1) and Sentinel-2 (12 bands, S2) pairs with 19 CORINE land-cover multi-labels; 11{,}310 train / 3{,}522 test.
\textbf{CBRSIR-VS}~\cite{cbrsirvs} pairs SAR (S1) and RGB (S2) views of the same scene, which differ strongly in appearance: 26{,}901 pairs, 10 classes, per-class 70/30 split (18{,}831 / 8{,}070).
\textbf{DSRSID}~\cite{li2018dsrsid} pairs Gaofen-1 high-resolution panchromatic (S1) and lower-resolution 4-band multispectral (S2) images: 80{,}000 pairs, 8 classes, per-class 70/30 split (56{,}000 / 24{,}000).
All images are resized to $120\times120$ and normalised per channel; every test sample is both query and gallery item.

%% file: sec/5_results.tex
\section{Results and Discussion}
\label{sec:results}
\vspace{-0.1em}

\paragraph{Quantitative Analysis.}
As shown in \cref{tab:main}, RS-JEPA achieves the best F1@5 in every direction on all three benchmarks, improving the mean over the strongest baseline by 3.11 points on BEN-14K (CR-JEPA), 2.28 on CBRSIR-VS (DeCUR) and 5.99 on DSRSID (SkySense).
The baselines rank inconsistently across datasets: DeCUR, for instance, leads on CBRSIR-VS but is among the weakest on DSRSID. RS-JEPA ranks first throughout.
Its advantage is consistently larger in cross-sensor than in same-sensor retrieval, in line with an objective that targets content shared by both sensors.
When each query's co-registered partner is removed from the cross-sensor gallery, RS-JEPA still leads on all benchmarks (70.46, 78.82 and 93.55 \vs 67.82, 75.07 and 85.81 for the strongest baselines), so its margin does not rely on instance matching.
Over the depths in \cref{tab:multik}, RS-JEPA is best in 59 of the 60 direction and average columns, the exception being S2$\to$S2 F1@20 on CBRSIR-VS, where DeCUR leads by 0.29 points. On DSRSID, its lead in average F1 widens from 2.88 points at $k=1$ to 7.36 at $k=20$, showing that its ranking stays reliable deeper into the retrieved list.
\vspace{-0.5 cm}

\paragraph{Qualitative Analysis.}
\Cref{fig:qualitative} shows the top-5 results for one query per direction and benchmark, selected by a fixed rule rather than by hand.
Over the two queries per benchmark, RS-JEPA retrieves at least as many relevant scenes as every other method, and all ten of its results on DSRSID match the query class.
A few queries cannot by themselves support general conclusions.
\vspace{-0.5 cm}

\begin{table}[!t]
\centering
\small
\caption{\textbf{Ablations.} Mean F1@5 and mAP (\%) over the four directions. \emph{Top:} components added cumulatively; \emph{bottom:} masking ratio. Best in \textbf{bold}, second \underline{underlined}.}
\label{tab:core_ablation}
\setlength{\tabcolsep}{2.5pt}
\resizebox{\columnwidth}{!}{%
\input{tables/table2_core_ablation}
}
\end{table}

\begin{figure}[!t]
\centering
\includegraphics[width=0.89\columnwidth]{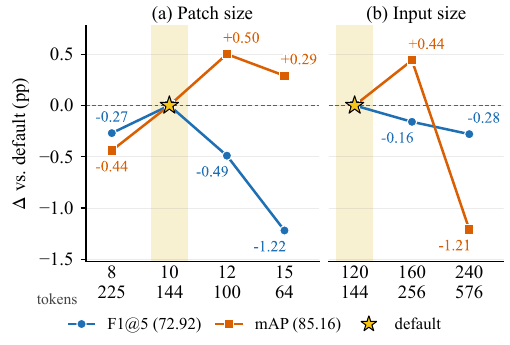}\\[4pt]
{\footnotesize (c) Original \vs our input setting}\\[2pt]
{\small\setlength{\tabcolsep}{4pt}\input{tables/table5_baseline_resolution}}
\caption{\textbf{Resolution} (BEN-14K). (a)~Patch size at $120\times120$, (b)~input size at $10\times10$ patches: change from our default (star), with token counts below. (c)~Mean F1@5 and mAP (\%) at each method's original and our setting.}
\label{fig:resolution}
\end{figure}

\begin{table}[!t]
\centering
\small
\caption{\textbf{Cross-sensor alignment.} mAP (\%) for same- and cross-sensor retrieval; Asym.: $|$S1$\to$S2 $-$ S2$\to$S1$|$. Best in \textbf{bold}.}
\label{tab:alignment}
\setlength{\tabcolsep}{2.5pt}
\resizebox{\columnwidth}{!}{%
\input{tables/table4_alignment}
}
\end{table}

\begin{table*}[!t]
\centering
\scriptsize
\setlength{\tabcolsep}{3.2pt}\renewcommand{\arraystretch}{0.78}
\caption{\textbf{Retrieval depth.} F1@$k$ (\%) for $k\in\{1,10,20\}$ in every direction (F1@5 in \cref{tab:main}); $i{\to}j$: query from sensor S$i$, gallery from sensor S$j$. Best in \textbf{bold}, second best \underline{underlined}, per column among the models at the same $k$. Cross-sensor F1@1 is high because the gallery contains the query's co-registered partner (\cref{sec:protocol}).}
\label{tab:multik}
\input{tables/table3_multik}

\end{table*}

\paragraph{Computational Efficiency Analysis.}
RS-JEPA uses 14.57M parameters during training, of which only 3.88M are retained for retrieval, since the AVC modules, predictors and target pass serve training alone.
Embedding one S1--S2 pair requires 2.23 GFLOPs, against 230.47M parameters and 32.98 GFLOPs for X-JEPA (\cref{tab:main}), corresponding to roughly 16$\times$ fewer parameters and 15$\times$ fewer operations.
\TANI{We also timed every model of \cref{tab:main} on one V100 with FP16. At batch size 64, RS-JEPA embeds 1{,}918 pairs per second, faster than all baselines except DeCUR (3{,}235), and needs the least GPU memory at batch 256 (0.60\,GiB against 1.34--3.90\,GiB).
For a single pair, however, it takes 33.1\,ms against 13.7--21.1\,ms for most baselines, as its many small scan kernels leave the GPU idle, so its advantage lies in batched embedding.}

%% file: tables/table2_core_ablation.tex
\begin{tabular}{@{}lcccccc@{}}
\toprule
 & \multicolumn{2}{c}{BEN-14K} & \multicolumn{2}{c}{CBRSIR-VS} & \multicolumn{2}{c}{DSRSID} \\
\cmidrule(lr){2-3} \cmidrule(lr){4-5} \cmidrule(lr){6-7}
Configuration & F1@5 & mAP & F1@5 & mAP & F1@5 & mAP \\
\midrule
\multicolumn{7}{@{}l}{\emph{Components}} \\
Base & 71.47$_{\pm 0.18}$ & 83.85$_{\pm 0.22}$ & 78.25$_{\pm 0.21}$ & 51.40$_{\pm 0.28}$ & 92.84$_{\pm 0.16}$ & 38.73$_{\pm 0.32}$ \\
$+$ Consensus & 71.82$_{\pm 0.15}$ & 84.42$_{\pm 0.19}$ & 79.18$_{\pm 0.19}$ & 53.19$_{\pm 0.25}$ & 93.58$_{\pm 0.14}$ & 42.93$_{\pm 0.29}$ \\
$+$ Scan views & \underline{72.64}$_{\pm 0.14}$ & \underline{84.85}$_{\pm 0.20}$ & \underline{79.82}$_{\pm 0.17}$ & \underline{54.43}$_{\pm 0.22}$ & \underline{94.06}$_{\pm 0.12}$ & \underline{45.83}$_{\pm 0.25}$ \\
$+$ AVC (Ours) & \textbf{72.92}$_{\pm 0.12}$ & \textbf{85.16}$_{\pm 0.18}$ & \textbf{80.35}$_{\pm 0.15}$ & \textbf{55.05}$_{\pm 0.20}$ & \textbf{94.43}$_{\pm 0.11}$ & \textbf{47.09}$_{\pm 0.22}$ \\
\midrule
\multicolumn{7}{@{}l}{\emph{Masking ratio (full model)}} \\
50\% & \underline{72.75}$_{\pm 0.16}$ & 84.33$_{\pm 0.21}$ & \underline{80.12}$_{\pm 0.18}$ & \textbf{55.33}$_{\pm 0.24}$ & 94.06$_{\pm 0.15}$ & 46.83$_{\pm 0.27}$ \\
60\% & \textbf{72.92}$_{\pm 0.12}$ & \underline{85.16}$_{\pm 0.18}$ & \textbf{80.35}$_{\pm 0.15}$ & \underline{55.05}$_{\pm 0.20}$ & \textbf{94.43}$_{\pm 0.11}$ & \underline{47.09}$_{\pm 0.22}$ \\
70\% & 72.49$_{\pm 0.17}$ & \textbf{86.40}$_{\pm 0.24}$ & 78.73$_{\pm 0.23}$ & 54.78$_{\pm 0.26}$ & \underline{94.22}$_{\pm 0.13}$ & \textbf{47.44}$_{\pm 0.24}$ \\
\bottomrule
\end{tabular}

%% file: tables/table5_baseline_resolution.tex
\begin{tabular}{@{}lcccc@{}}
\toprule
 & \multicolumn{2}{c}{$224{\times}224$, $16{\times}16$} & \multicolumn{2}{c}{$120{\times}120$, $10{\times}10$} \\
\cmidrule(lr){2-3} \cmidrule(l){4-5}
Method & F1@5 & mAP & F1@5 & mAP \\
\midrule
X-JEPA~\cite{xjepa} & 68.47 & 81.49 & 67.79 & 81.30 \\
CR-JEPA~\cite{crjepa} & 69.86 & 80.26 & 69.81 & 81.08 \\
\rowcolor{oursbg}\textbf{RS-JEPA (Ours)} & \textbf{72.38} & \textbf{85.40} & \textbf{72.92} & \textbf{85.16} \\
\bottomrule
\end{tabular}

%% file: tables/table4_alignment.tex
\begin{tabular}{@{}l>{\hspace{3pt}}ccc>{\hspace{3pt}}ccc>{\hspace{3pt}}ccc@{}}
\toprule
\multirow{2}{*}{Model} & \multicolumn{3}{c}{BEN-14K} & \multicolumn{3}{c}{CBRSIR-VS} & \multicolumn{3}{c}{DSRSID} \\
\cmidrule(lr){2-4} \cmidrule(lr){5-7} \cmidrule(l){8-10}
 & Same & Cross & Asym.$\downarrow$ & Same & Cross & Asym.$\downarrow$ & Same & Cross & Asym.$\downarrow$ \\
\midrule
CROMA & \textbf{87.02} & \textbf{86.84} & 0.16 & 49.82 & 48.81 & 0.50 & 38.87 & 38.50 & 0.04 \\
DeCUR & 83.21 & 83.06 & 1.03 & 47.83 & 44.71 & 1.58 & 29.56 & 27.24 & 0.14 \\
SkySense & 81.45 & 80.40 & 0.16 & 34.99 & 31.78 & 0.10 & 46.17 & 45.60 & 0.10 \\
X-JEPA & 81.36 & 81.23 & 0.89 & 48.52 & 47.89 & 2.50 & 33.54 & 33.57 & 0.04 \\
CR-JEPA & 81.87 & 80.28 & 0.28 & 43.97 & 47.73 & 0.50 & 30.23 & 28.19 & 0.09 \\
\midrule
\rowcolor{oursbg}\textbf{RS-JEPA (ours)} & 85.26 & 85.06 & \textbf{0.12} & \textbf{55.42} & \textbf{54.68} & \textbf{0.08} & \textbf{47.35} & \textbf{46.82} & \textbf{0.03} \\
\bottomrule
\end{tabular}

%% file: tables/table3_multik.tex
\begin{tabular}{@{}lc>{\hspace{4pt}}ccccc>{\hspace{4pt}}ccccc>{\hspace{4pt}}ccccc@{}}
\toprule
\multirow{2}{*}{Method} & \multirow{2}{*}{$k$} & \multicolumn{5}{c}{BEN-14K} & \multicolumn{5}{c}{CBRSIR-VS} & \multicolumn{5}{c}{DSRSID} \\
\cmidrule(lr){3-7} \cmidrule(lr){8-12} \cmidrule(l){13-17}
 & & 1$\to$1 & 2$\to$2 & 1$\to$2 & 2$\to$1 & Avg. & 1$\to$1 & 2$\to$2 & 1$\to$2 & 2$\to$1 & Avg. & 1$\to$1 & 2$\to$2 & 1$\to$2 & 2$\to$1 & Avg. \\
\midrule
 & 1 & 66.48 & 70.70 & 74.28 & 74.43 & 71.47 & 65.70 & 75.39 & 65.18 & 66.56 & 68.21 & 77.38 & 75.89 & \underline{98.43} & \underline{98.20} & 87.48 \\
 & 10 & 64.26 & 67.81 & 66.62 & 65.66 & 66.09 & 63.35 & 71.91 & 62.58 & 63.70 & 65.39 & 73.46 & 71.49 & 73.73 & 73.23 & 72.98 \\
\multirow{-3}{*}{MAE-RVSA~\cite{wang2023maervsa}} & 20 & 63.31 & 66.60 & 64.67 & 63.85 & 64.61 & 62.01 & 70.29 & 61.19 & 62.23 & 63.93 & 71.67 & 69.57 & 70.65 & 70.10 & 70.50 \\
\cmidrule(l){1-17}
 & 1 & 64.94 & 70.14 & \underline{76.08} & \underline{77.70} & \underline{72.22} & 66.77 & 77.21 & 66.25 & 67.43 & 69.41 & 75.96 & 74.53 & 97.41 & 96.92 & 86.20 \\
 & 10 & 63.10 & 68.00 & 65.67 & 66.47 & 65.81 & 63.94 & 73.89 & 63.53 & 64.94 & 66.58 & 72.03 & 70.34 & 71.74 & 71.10 & 71.30 \\
\multirow{-3}{*}{ScaleMAE~\cite{reed2023scalemae}} & 20 & 62.25 & 66.83 & 63.87 & 64.44 & 64.35 & 62.57 & 72.10 & 62.06 & 63.44 & 65.04 & 70.22 & 68.48 & 68.40 & 67.63 & 68.68 \\
\cmidrule(l){1-17}
 & 1 & 66.89 & 70.36 & 73.12 & 74.44 & 71.20 & 66.73 & 76.90 & 66.85 & 66.37 & 69.21 & 79.42 & 79.57 & 93.95 & 95.00 & 86.99 \\
 & 10 & 63.73 & 68.15 & 65.58 & 66.08 & 65.88 & 63.78 & 73.03 & 64.47 & 64.57 & 66.46 & 75.63 & 75.53 & 74.76 & 75.00 & 75.23 \\
\multirow{-3}{*}{SatMAE++~\cite{noman2024satmaepp}} & 20 & 62.58 & 66.95 & 63.98 & 64.15 & 64.42 & 62.36 & 71.17 & 62.99 & 63.11 & 64.91 & 73.97 & 73.71 & 72.01 & 72.18 & 72.97 \\
\midrule
 & 1 & \underline{69.45} & 72.30 & 69.91 & 71.34 & 70.75 & 73.68 & 81.82 & 72.32 & 74.19 & 75.50 & 93.03 & 93.37 & 89.86 & 89.45 & 91.42 \\
 & 10 & \underline{67.86} & \underline{70.54} & \underline{68.18} & 69.10 & \underline{68.92} & 71.42 & 79.08 & 71.72 & 72.76 & 73.74 & 85.64 & \underline{86.44} & 85.61 & 85.10 & 85.70 \\
\multirow{-3}{*}{CROMA~\cite{fuller2023croma}} & 20 & \underline{66.95} & \underline{69.61} & \underline{67.29} & \underline{68.04} & \underline{67.97} & 70.37 & 77.73 & 70.89 & 71.73 & 72.68 & \underline{83.97} & \underline{84.71} & \underline{83.79} & \underline{83.28} & \underline{83.93} \\
\cmidrule(l){1-17}
 & 1 & 66.89 & 67.74 & 64.24 & 62.92 & 65.45 & \underline{77.05} & \underline{86.77} & \underline{72.86} & \underline{77.37} & \underline{78.51} & 80.70 & 78.18 & 63.43 & 61.27 & 70.90 \\
 & 10 & 65.52 & 65.88 & 61.93 & 61.29 & 63.66 & \underline{75.48} & \underline{85.29} & \underline{72.49} & \underline{76.97} & \underline{77.56} & 71.96 & 71.92 & 61.92 & 61.16 & 66.74 \\
\multirow{-3}{*}{DeCUR~\cite{wang2024decur}} & 20 & 64.86 & 65.27 & 61.40 & 60.99 & 63.13 & \underline{74.77} & \textbf{84.45} & \underline{72.08} & \underline{75.82} & \underline{76.78} & 70.65 & 70.64 & 61.29 & 60.83 & 65.85 \\
\cmidrule(l){1-17}
 & 1 & 67.63 & 72.34 & 70.34 & 70.83 & 70.29 & 67.40 & 76.72 & 67.00 & 66.18 & 69.32 & 93.35 & 93.23 & 96.82 & 96.88 & \underline{95.07} \\
 & 10 & 65.84 & 70.10 & 66.87 & 66.76 & 67.39 & 65.43 & 72.93 & 64.56 & 63.92 & 66.71 & \underline{85.82} & 85.99 & \underline{85.95} & \underline{85.70} & \underline{85.86} \\
\multirow{-3}{*}{SkySense~\cite{guo2024skysense}} & 20 & 64.85 & 68.88 & 65.54 & 65.47 & 66.18 & 64.20 & 71.01 & 63.17 & 62.82 & 65.30 & 83.52 & 83.84 & 83.30 & 83.04 & 83.43 \\
\midrule
 & 1 & 67.15 & 70.81 & 68.32 & 69.89 & 69.04 & 68.27 & 83.10 & 72.59 & 71.92 & 73.97 & \underline{94.38} & \underline{93.94} & 94.11 & 93.89 & 94.08 \\
 & 10 & 65.42 & 69.10 & 66.22 & 67.13 & 66.97 & 67.40 & 80.68 & 71.36 & 71.30 & 72.68 & 83.28 & 83.49 & 84.66 & 84.72 & 84.03 \\
\multirow{-3}{*}{X-JEPA~\cite{xjepa}} & 20 & 64.57 & 67.99 & 65.42 & 66.12 & 66.02 & 67.18 & 79.71 & 70.92 & 70.77 & 72.14 & 81.23 & 81.47 & 82.11 & 82.18 & 81.75 \\
\cmidrule(l){1-17}
 & 1 & 68.76 & \underline{73.42} & 72.39 & 73.25 & 71.95 & 61.52 & 77.10 & 61.91 & 65.63 & 66.54 & 72.79 & 75.37 & 86.83 & 87.32 & 80.58 \\
 & 10 & 67.12 & 70.51 & 68.03 & \underline{69.15} & 68.70 & 60.18 & 73.34 & 61.65 & 64.14 & 64.83 & 70.59 & 72.83 & 72.08 & 72.56 & 72.02 \\
\multirow{-3}{*}{CR-JEPA~\cite{crjepa}} & 20 & 66.15 & 69.33 & 66.88 & 67.78 & 67.53 & 59.60 & 71.84 & 61.51 & 63.44 & 64.10 & 69.58 & 71.72 & 70.11 & 70.63 & 70.51 \\
\midrule
\rowcolor{oursbg} & 1 & \textbf{72.25} & \textbf{74.32} & \textbf{81.95} & \textbf{82.39} & \textbf{77.73} & \textbf{77.43} & \textbf{88.45} & \textbf{79.47} & \textbf{80.81} & \textbf{81.54} & \textbf{96.09} & \textbf{95.81} & \textbf{99.99} & \textbf{99.91} & \textbf{97.95} \\
\rowcolor{oursbg} & 10 & \textbf{69.95} & \textbf{71.48} & \textbf{71.71} & \textbf{72.06} & \textbf{71.30} & \textbf{75.81} & \textbf{85.64} & \textbf{78.24} & \textbf{78.77} & \textbf{79.61} & \textbf{92.54} & \textbf{92.52} & \textbf{93.35} & \textbf{93.36} & \textbf{92.94} \\
\rowcolor{oursbg}\multirow{-3}{*}{\textbf{RS-JEPA (Ours)}} & 20 & \textbf{68.85} & \textbf{70.36} & \textbf{69.95} & \textbf{70.27} & \textbf{69.86} & \textbf{74.99} & \underline{84.16} & \textbf{77.56} & \textbf{77.71} & \textbf{78.61} & \textbf{91.10} & \textbf{91.03} & \textbf{91.50} & \textbf{91.51} & \textbf{91.29} \\
\bottomrule
\end{tabular}
\vspace{-4mm}

%% file: sec/6_ablation.tex
\section{Ablation Studies}
\label{sec:ablation}
All defaults (60\% masking, $10\times10$ patches, $120\times120$ input) were fixed \emph{a priori} from prior JEPA work; the ablations below are post-hoc sensitivity analyses.

\textbf{Components:}
Starting from a single-view base (row-major scan; $\mathcal{L}_{\mathrm{ctx}}$, $\mathcal{L}_{\mathrm{nce}}$, $\mathcal{L}_{\cos}$), \cref{tab:core_ablation} (top) adds consensus ($\mathcal{L}_{\mathrm{rel}}$), then scan views (random paths and $\mathcal{L}_{\mathrm{view}}$ from $\mathbf{C}$, no AVC), then AVC. Each step improves F1@5 and mAP on all three benchmarks, and the full model is best (72.92/85.16 on BEN-14K, 80.35/55.05 on CBRSIR-VS, 94.43/47.09 on DSRSID).

\textbf{Component Variants:}
On BEN-14K, removing consensus changes F1@5 by only 0.10 but lowers mAP by 3.20; a symmetric cyclic loss in its place lowers F1@5 by 0.61 and raises mAP by 1.57. A random second path matches \cref{eq:g2}. \TANI{We did not run random splits, AVC without coupled attention, or a fixed-path control with the same $\mathcal{L}_{\mathrm{view}}$, so \cref{tab:core_ablation} cannot separate adversarial partitioning from coupled attention, nor scan order from the extra prediction loss.}

\textbf{Masking Ratio:}
From 50\% to 70\% masking (\cref{tab:core_ablation}, bottom), F1@5 varies by at most 0.43 on BEN-14K and 0.37 on DSRSID, but drops by 1.62 on CBRSIR-VS at 70\%, with lower mAP, suggesting heavy masking suits its large SAR--RGB appearance gap poorly.

\textbf{Patch and Input Size:}
On BEN-14K (\cref{fig:resolution}a,b; larger inputs upsampled), F1@5 peaks at the default (72.92) and stays within 1.22 points elsewhere, while mAP gains at most 0.50 with $12\times12$ or $15\times15$ patches or $160\times160$ inputs. Smaller patches and larger inputs need 1.6--4$\times$ more tokens without improving F1@5.

\textbf{Cross-Sensor Alignment:}
In full-ranking mAP (\cref{tab:alignment}), RS-JEPA is best for same- and cross-sensor retrieval on CBRSIR-VS (55.42/54.68) and DSRSID (47.35/46.82) and second to CROMA on BEN-14K. Its asymmetry between the two cross-sensor directions is the lowest on all benchmarks (0.12, 0.08, 0.03), so retrieval is equally reliable whichever sensor issues the query.

%% file: sec/7_conclusion.tex
\section{Conclusion}
\label{sec:conclusion} 
We proposed RS-JEPA, a compact JEPA for label-free cross-sensor retrieval that turns the reading order of a shared Vision Mamba encoder into scan-path views, makes their prediction harder with Adversarial View Coupling, and aligns the sensors through neighbourhood consensus. With a 3.88M-parameter retrieval network, RS-JEPA achieves the highest F1@5 of twelve re-implemented methods across all three benchmarks, while CROMA remains ahead in full-ranking mAP on BEN-14K. \TANI{\textbf{Limitations.} RS-JEPA still needs co-registered training pairs, limiting scale. We test three benchmarks at $120\times120$ pixels, with no unseen regions or other sensors, one fixed scan-path set and a shared baseline budget. Ablations are scored on the test splits, so they show trends only. Training adds 10.7M parameters; its time cost is unmeasured. Broader archives and transfer to new regions are future work.} Reading order is thus an effective source of self-supervised views.